\documentclass[letterpaper, 10 pt, conference]{ieeeconf}  
\usepackage{graphicx}
\usepackage{subfig}
\usepackage{multirow}
\usepackage{float}
\usepackage{balance}

\IEEEoverridecommandlockouts                              
  
\title{\LARGE \bf
	Automatic Model Based Dataset Generation for Fast and Accurate Crop and Weeds Detection
}

\author{Maurilio Di Cicco, Ciro Potena, Giorgio Grisetti and Alberto Pretto
	\thanks{This work has been supported by the European Commission under the grant number H2020-ICT-644227-FLOURISH. Di Cicco, Potena, Grisetti and Pretto are with the Department of Computer, Control, and Management Engineering ``Antonio Ruberti``, Sapienza University of Rome, Italy. Email: {\tt \{dicicco, potena, grisetti, pretto\}@diag.uniroma1.it}.}.}

\begin{document}
	
	\maketitle
	\thispagestyle{empty}
	\pagestyle{empty}
	
	\begin{abstract}
Selective weeding is one of the key challenges in the field of agriculture robotics. To accomplish this task, a farm robot should be able to accurately detect plants and to distinguish them between crop and weeds. Most of the promising state-of-the-art approaches make use of appearance-based models trained on large annotated datasets. Unfortunately, creating large agricultural datasets with pixel-level annotations is an extremely time consuming task, actually penalizing the usage of data-driven techniques.
		
In this paper, we face this problem by proposing a novel and effective approach that aims to dramatically minimize the human intervention needed to train the detection and classification algorithms.			
The idea is to procedurally generate large synthetic training datasets randomizing the key features of the target environment (i.e., crop and weed species, type of soil, light conditions).
More specifically, by tuning these model parameters, and exploiting a few real-world textures, it is possible to render a large amount of realistic views of an artificial agricultural scenario with no effort.

The generated data can be directly used to train the model or to supplement real-world images.
We validate the proposed methodology by using as testbed a modern deep learning based image segmentation architecture.
We compare the classification results obtained using both real and synthetic images as training data.
The reported results confirm the effectiveness and the potentiality of our approach.	

\end{abstract}

\section{Introduction}

\begin{figure}
	\centering
	\includegraphics[width=0.9\columnwidth]{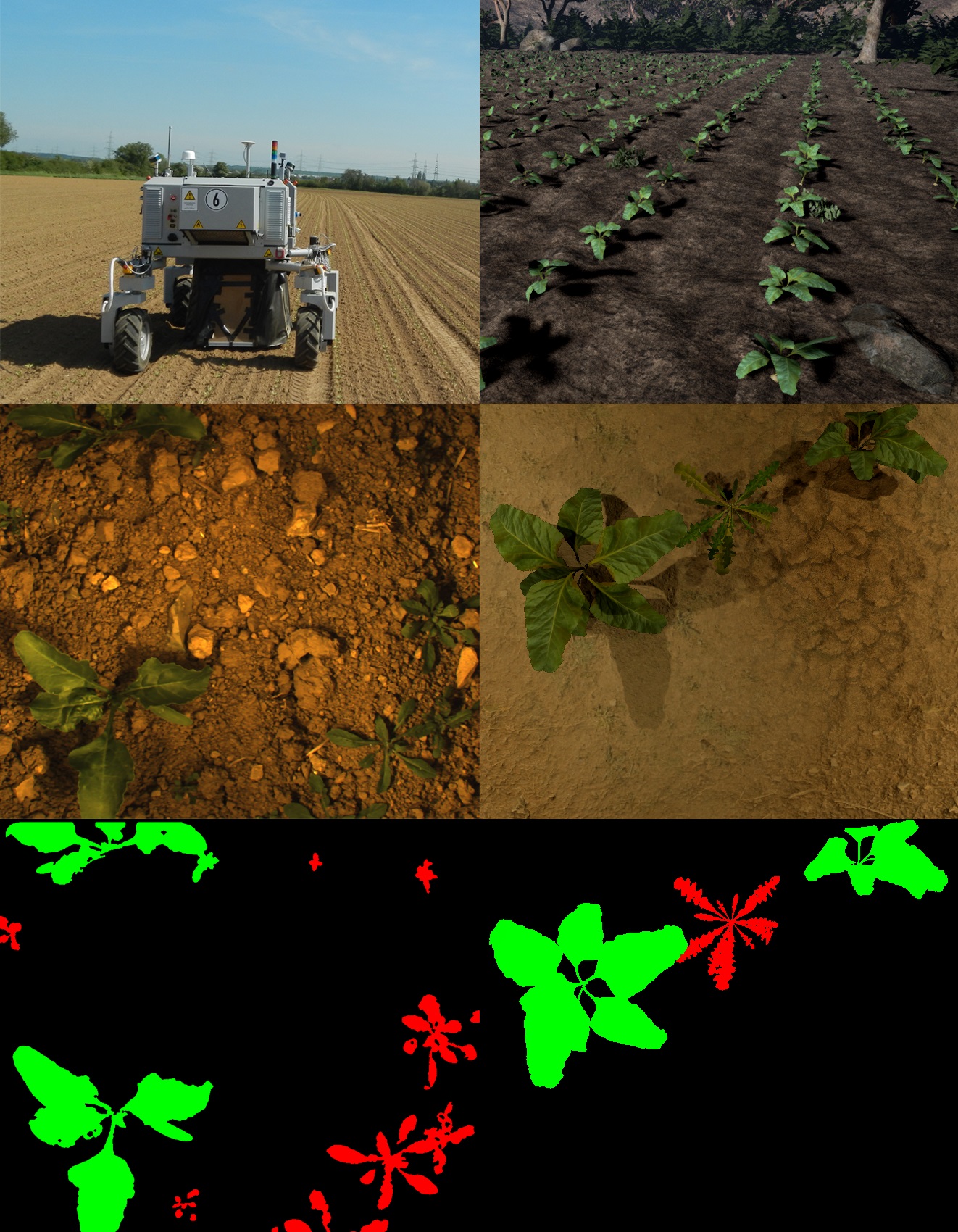}
	\caption{\small On the top left, BOSCH BoniRob employed to acquire the datasets used in the experiments, on top right an artificial sugar beet field generated using our procedure.\\
		On the second and third rows, on the left side real world images of sugar beets and \textit{Capsella Bursa-Pastoris} with its own hand-made ground truth. On the right side the specular synthetic generated one.}
	\label{fig:teaser}
\end{figure}     

Driven by a growing demand for a sustainable and efficient agriculture, in the last years many researchers focused on the precision farming domain.
In this area, one of the most important goal is to improve the farm productivity while reducing the usage of herbicides and pesticides. Precision farming applications address this challenge by means of cyclical and exhaustive field measurements such as plant's health indicators and density of weeds. Unfortunately, exhaustive and periodical data collection over the entire field is a time consuming and expensive activity.\\
The application of autonomous robots in the agricultural context can lead to a strong minimization of the human effort required in several agricultural tasks. In this paper we focus on the crop and weeds detection task, that represents one of the most challenging problems that should be addressed to build effective farming robots. 
The main goal is to develop an image-based classification system capable to identify the crop, distinguishing it from weeds. An illustration of an image-based classification input and output is reported in Fig. \ref{fig:teaser}(more specifically, second and third rows on the left column).

The most promising state-of-the-art approaches in this area usually make use of machine learning techniques, such as Convolutional Neural Networks (CNNs) (\cite{Potena_2016, reyes_clef2015, krizhevsky_nips2012}) or random forest classifiers \cite{lottes_icra2016}. 
    The usage of such techniques, especially CNNs, allows to train highly discriminative visual models capable to distinguish among different plant species with great accuracy.
The major drawback of these data-driven approaches is that the level of expressivity is limited by the size of the training dataset.
In the context of precision agriculture, the requirement for large datasets leads to a significant effort. More specifically, datasets should be acquired across different plant growth stages and weather conditions. In addition, the training images must be provided with an accurate semantic pixel-wise annotation.
Unfortunately, manual pixel-wise annotation is a challenging and extremely time consuming task. 
Actually, due to this problem, the size of the semantic datasets is usually relatively small \cite{Xie_2016}.\\
In this work, we explore the use of an open-source graphic engine as a solution for the above-mentioned problem, i.e. by creating data algorithmically. 
What makes this problem challenging is the realism and the fidelity required to reproduce the key aspects of the target environment, i.e. the virtual scenario must resemble as close as possible to the real one. This requires a precise modeling in terms of texture, 3D models and light conditions. 
Previous work on virtual dataset creation have focused mainly on handcrafted virtual worlds, moving the human effort from the annotation process to the synthetic dataset creation (e.g., \cite{Mancini_2016,Hattori_2015}).
Unlike these approaches, in this paper we focus on the procedural generation of virtual datasets, allowing us to potentially create an infinite number of synthetic images without any manual labor. More specifically, we parametrize the target environment with a set of key rules. We generate each synthetic scene by using few real world textures (e.g., plant leaf textures, soil textures, \dots) and by modulating the chosen environmental parameters (e.g., weather conditions, size of plants, \dots).\\
We compare the quality\footnote{The term ''quality`` denotes here the amount and quality of the information transfered to the trained model.} of the synthetic datasets using them to train a modern deep learning architecture, and then testing it on real data.
As the results suggest, even if the virtual scene does not contain all the weed species, the level of accuracy approximates the accuracy reached with real data. 
We also preformed the same tests training the testbed network with a synthetic dataset augmented with a small number of real images, obtaining even better results. Here the idea was to emulate a real use-case, where only a limited amount of annotated data is available. 
    
	\subsection{Related Work}
	
	\subsubsection{Plants Classification}
	The problem of plant classification can be considered an instance of the so-called \textit{fine-grained visual classification} (FGVC) problem, where the purpose is to recognize subordinate categories such as species of animals, models of cars, etc.  FGVC problems are intrinsically difficult since the differences between similar categories (in our case, plant species) are often minimal, and only in recent works the researchers obtained noteworthy results (e.g., \cite{parkhi12cat,ykf2011}).\\
	
	Burks \textit{et al.} \cite{Burks2000} proposed to use CCM texture statistics as input variables for a backpropagation (BP) neural network for weed classification. Feyaerts and van Gool \cite{Feyaerts2001} presented a performance of a classifier based on multispectral reflectance in order to distinguish the crop from weeds. The best classifier, based on neural networks (NN), reached a classification rate of 80\% for sugar beet plants and 91\% for weeds. Also Aitkenhead \textit{et al.} \cite{Aitkenhead2003} proposed a NN based approach: the captured images were first segmented into cells that are successively classified, achieving a final accuracy up to 75\%. 
	
	In Haug \textit{et al.} \cite{Haug2014} a Random Forest (RF) classifier was proposed. It uses a large number of simple features extracted from a large overlapping neighborhood around sparse pixel positions. This approach achieves strong classification accuracies, due to its ability of discriminating also crops that are very similar to weeds. This approach has been improved in Lottes \textit{et al.} \cite{lottes_icra2016} by extending the features set and including a relative plant arrangement prior that helps to obtain better classification results.\\

	Recently Han Lee \textit{et al.} \cite{hanlee_icip2015} presented a leaf-based plant classification system that uses convolutional neural networks to automatically learn suitable visual features. Also Reyes \textit{et al.} \cite{reyes_clef2015} used CNN for fine-grained plant classification: they used a deep CNN with the architecture proposed by Krishevsky \textit{et al.} \cite{krizhevsky_nips2012}, first initialized to recognize 1000 categories of generic objects, then \textit{fine-tuned} (i.e., specialized) for the specific task to recognize 1000 possible plant species
	
	\subsubsection{Synthetic Dataset Generation}
	Modern data-driven classification approaches like CNN architectures require a large amount of data to obtain the best performance.
	One recently proposed solution to address this issue is to generate synthetically the training datasets. Many approaches addressed this issue by exploiting modern graphic engines.		
	\cite{Richter_2016, Shafei_2017} present approaches based on synthetic data extracted from computer video games, showing how a merged training dataset composed of synthetic and real world images performs better than a real one.
	Modern video games are a compelling source of high quality annotated data but they also have an important drawback. They are usually closed-source, so it is not possible to customize and modify the output data stream.
	In other approaches the artificial world is totally handcrafted using modern graphic tools. Mancini \textit{et al.} \cite{Mancini_2016} presented an urban scenario developed by means of Unreal Engine 4 for the monocular depth estimation. 
	Experiments using the synthetic data for the training phase, without any fine tuning, show good generalization properties on real data. A similar technique has been used in \cite{Hattori_2015} to build a city environment for pedestrian detection, showing how using purely synthetic data is able to outperform models trained on a limited amount of real world scenes. 
	In that kind of solution the great human effort required in the dataset acquisition and labeling is transfered into the synthetic world crafting. Even if the latter involves a minor effort, it is still a time consuming activity and the amount of  generated data depends on the size of the generated artificial world.\\
	
	Another approach used to mitigate the labeling effort is based on transfer learning \cite{Pan_2010}. For instance Yosinski et al. \cite{Yosinski_2014} show how better is the transferability of features depending on the distance between the base and the target task.
	
	A different approach is based on dataset reduction: in our previous work \cite{Potena_2016}, we started from a large dataset and by using a dataset summarization technique, we built the training dataset by choosing only a sub-optimal subset of images that maximizes the visual entropy. 
	
	\subsection{Contributions}	
	
	The main contributions of this work include: 
	\begin{itemize}
	  \item Unlike previous custom-built solutions, we explicitly model all the relevant aspects without manually creating the entire scene;
	  \item By modulating and randomizing the model parameters, we are able to quickly generate a large amount of scenes in a procedural fashion.\\
	\end{itemize}

	The remainder of this paper is organized as follows. Section II describes the whole environmental modeling and the automatic ground truth generation. In section III we show the results and, finally, in section IV we draw the conclusions. 
	
	\section{Procedural Dataset Generation}
	Procedural generation is a widely used technique in computer graphics, and it has been exploited in several scenarios, such as virtual cities generation \cite{greuter03} and virtual dungeons creation \cite{LLB14}.
	The goal of the procedural dataset generation is to build a randomized rendering pipeline. In our case, this can be viewed as a generative model from which we can sample fully labeled training images of agricultural scenes. The main goals in building this pipeline were twofold: (i) \textit{Realism}, the synthetic agricultural data has to closely resemble the real one, letting the trained model to work in real agricultural environment; (ii) \textit{variety}, the artificial dataset must guarantee a good coverage over the appearance variations, resulting in an unpredictable range of possible scenes.  
	In this section we describe how we take into account these issues in modeling a virtual agricultural scene. Firstly, we describe the generic kinematic model of a leaf prototype that we use to generate leaves of different plant species. Then we show how to assemble the single leaf meshes into the target artificial plant.
	Finally, we explain the scattering scheme with which we place the artificial plants on the realistic soil and the rendering procedure that allows to obtain at no cost annotated data from our virtual crop field.
	
	\subsection{Description of the Leaf Model}
	
	\begin{figure}
		\centering
		\includegraphics[width=0.6\columnwidth]{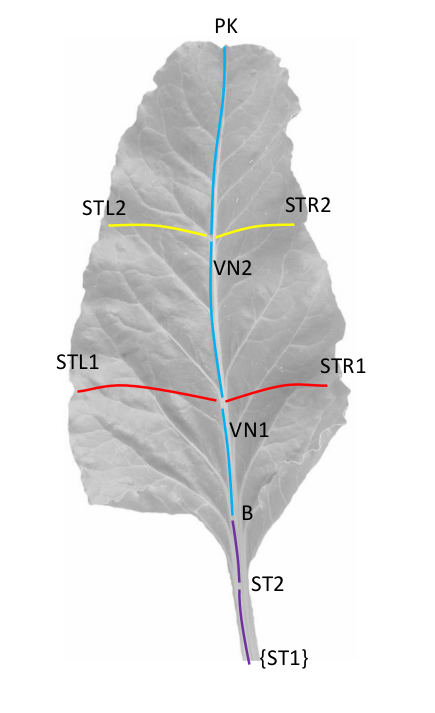} 
		\caption{Leaf kinematic model representing a leaf.}
		\label{fig:leaf_model}
	\end{figure}      
	
	\begin{figure}
		\centering
		\begin{tabular}{ccccc}
			\subfloat[Mesh]{\includegraphics[width=0.15\columnwidth]{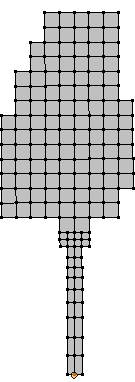}} & 
			\subfloat[Normal]{\includegraphics[width=0.15\columnwidth]{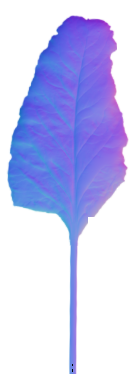}} & 
			\subfloat[AO]{\includegraphics[width=0.15\columnwidth]{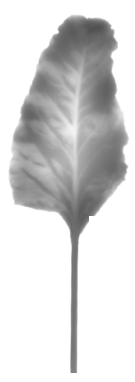}} &
			\subfloat[Height]{\includegraphics[width=0.15\columnwidth]{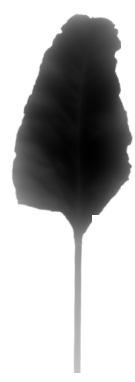}}& 
			\subfloat[Final]{\includegraphics[width=0.15\columnwidth]{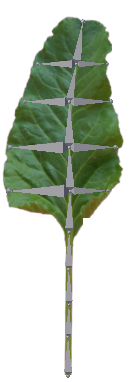}} 
		\end{tabular}
		\caption{\small (a) Shows the the generic planar surface mesh. The number of polygons' subdivision can be kept as low as possible at the expense of deformation quality while acting on the skeleton. (b) Highlights the Normal Map while (c) and (d) respectively represent the Ambient Occlusion and the Height map. (e) Shows the final shading results as well as the skeleton.
\label{fig:sugar_beet_leaf_process}
		}
	\end{figure}
	
	The general kinematic model of the simulated leaf is devised as a \textit{kinematic tree} where the root node is the first reference system of the stem, {ST1} (see Fig. \ref{fig:leaf_model}). The chain follows the stem direction until the base of the leaf (B), and then the leaf's principal vein, reaching finally the peak (PK). 
	Each joint onto the leaf (VNi) can be used as starting point for two mirrored branches (STLi, STRi) with respect to the leaf principal axis, following the secondary veins.      
	
	The posture of the j-th leaf joint with respect to the parent one is parametrized, according to the Denavit-Hartenberg convention, with a single rotation around the z axis (see Fig. \ref{fig:leaf_model}).      
	The benefit	of using such a method is twofold. Firstly, it allows to cover a wide variety of crop and weeds realistic leaves in different growth stages just choosing the number of vein joints and their relative distances.
	
	Secondly, acting on the kinematic chain angles as input parameters it is possible to bend the artificial leaf to resemble physical effects such as the gravity. In addition, and most importantly, adding a random component to such angles in the generation phase leads to an unpredictable range of possible leaves.
	 
	Once the leaf kinematic model is ready, the skeleton obtained from such model is associated to a planar surface mesh. 
	The artificial leaf model is then physically based shaded \cite{pharr2004physically} by means of a high definition RGBA texture taken from real world pictures. The alpha channel represents the opacity mask. 
	
		\begin{figure*}[!ht]
		\centering
		\begin{tabular}{cccc}
			{\includegraphics[width=0.18\textwidth]{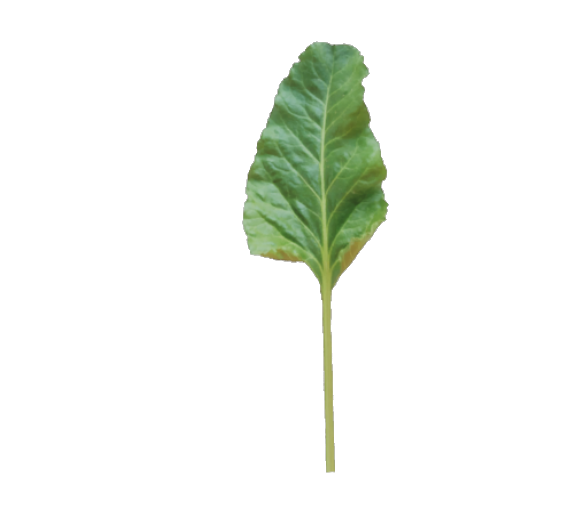}} & 
			{\includegraphics[width=0.18\textwidth]{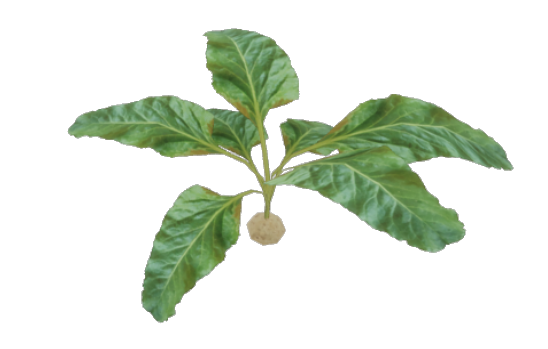}} &
			{\includegraphics[width=0.18\textwidth]{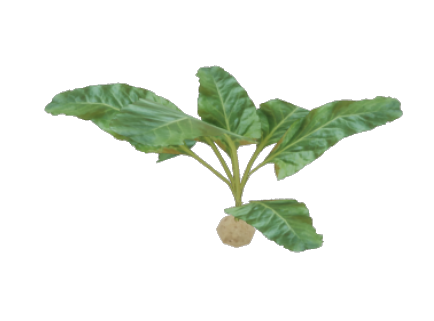}} &
			{\includegraphics[width=0.18\textwidth]{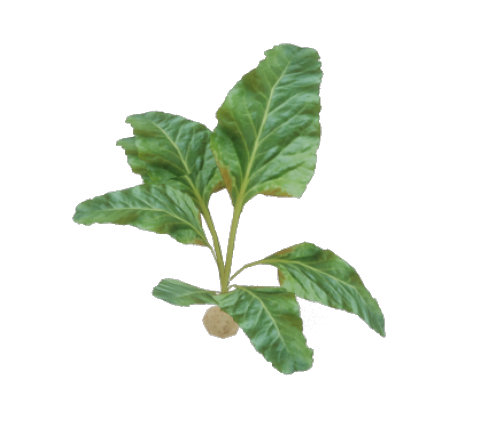}}\\
			{\includegraphics[width=0.18\textwidth]{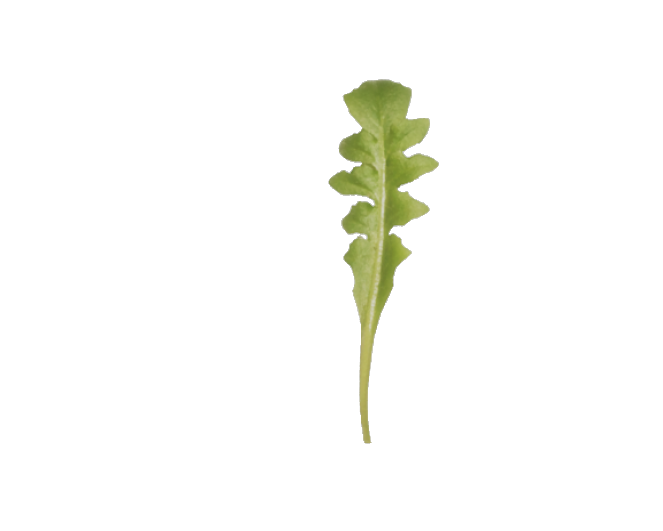}} & 
			{\includegraphics[width=0.18\textwidth]{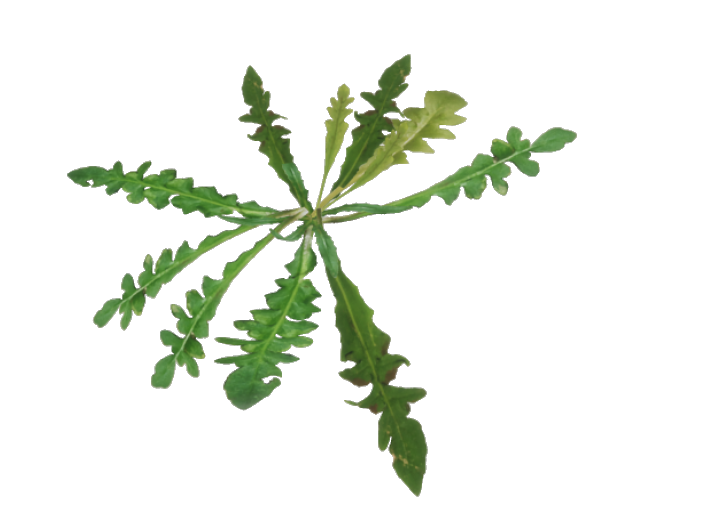}} &
			{\includegraphics[width=0.18\textwidth]{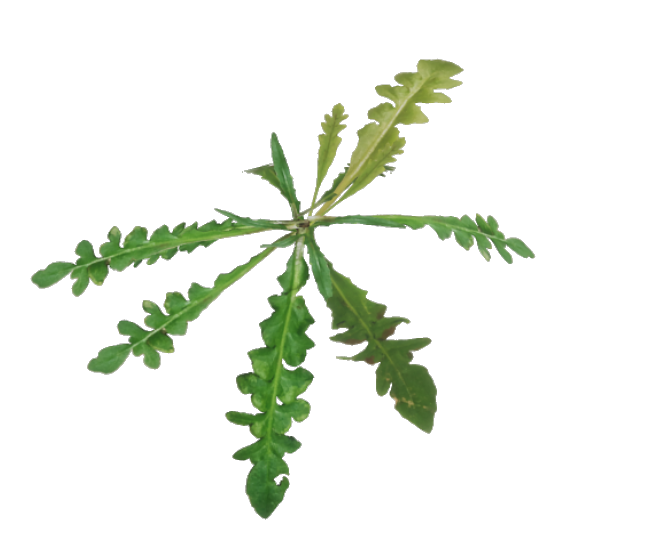}} &
			{\includegraphics[width=0.18\textwidth]{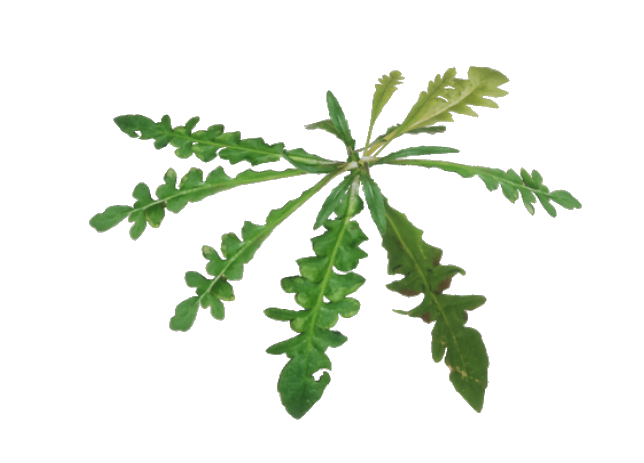}}\\
			{\includegraphics[width=0.18\textwidth]{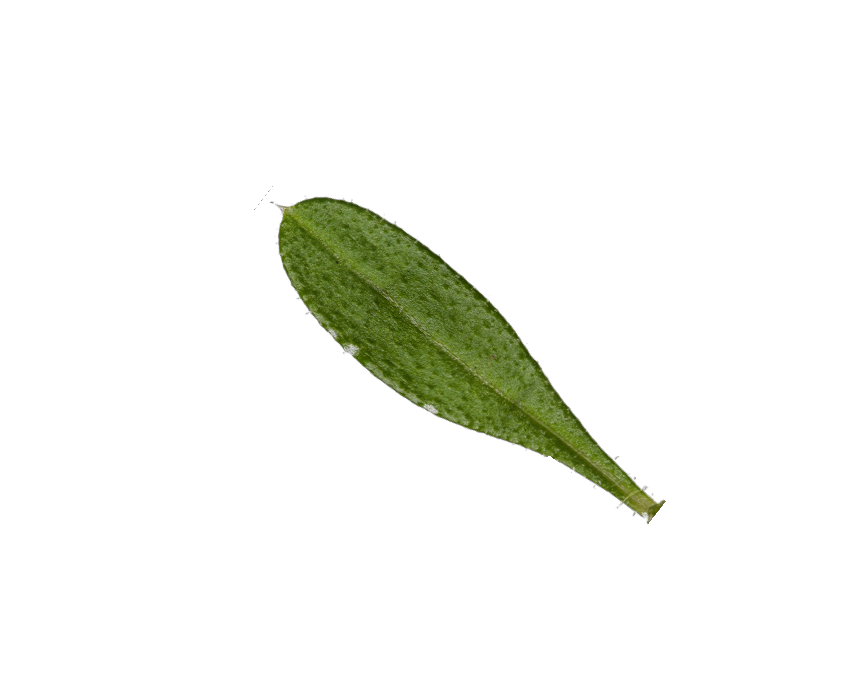}} & 
			{\includegraphics[width=0.18\textwidth]{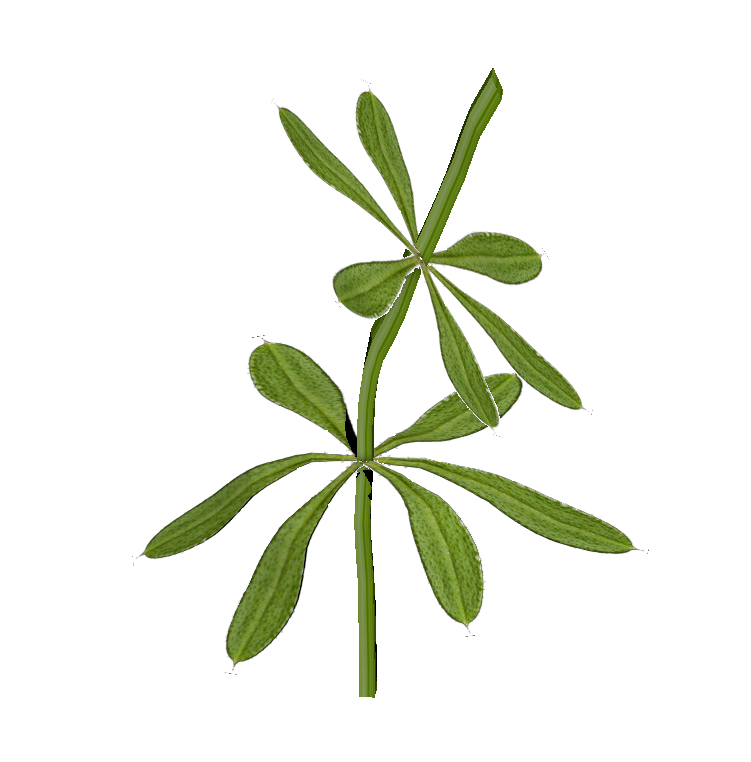}} &
			{\includegraphics[width=0.18\textwidth]{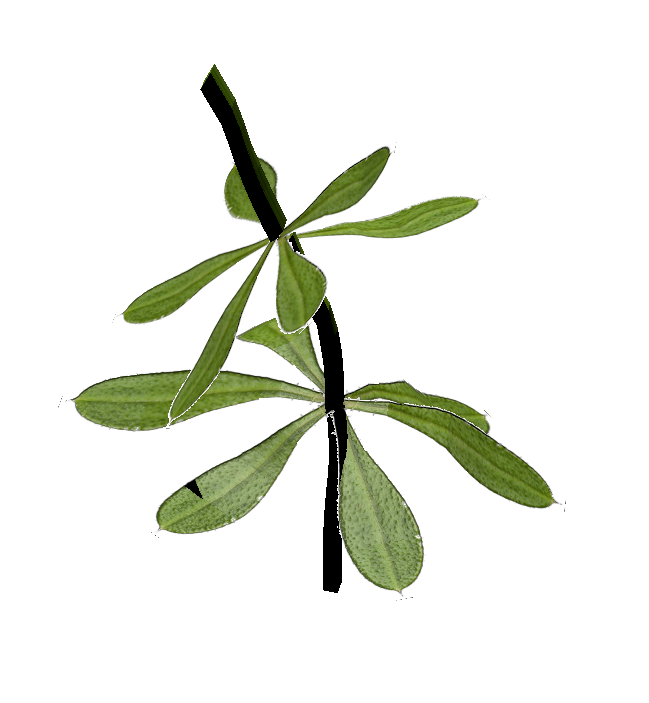}} &
			{\includegraphics[width=0.18\textwidth]{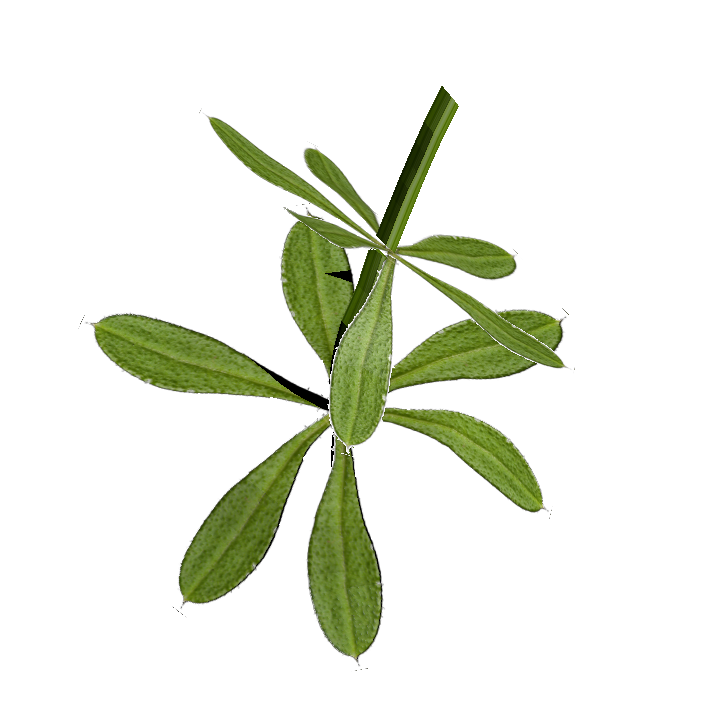}}\\            
		\end{tabular}
		
		\caption{\small Examples of three different weeds obtained from a single leaf texture. In the top row three instances of sugar beets, in the second row three different \textit{Capsella Bursa-Pastoris} weeds, in the last row three \textit{Galium Aparine} specimens.\\}
		\label{fig:plant_specimans}
	\end{figure*}	
	
		\begin{figure}[ht]
		\centering
\includegraphics[width=\columnwidth]{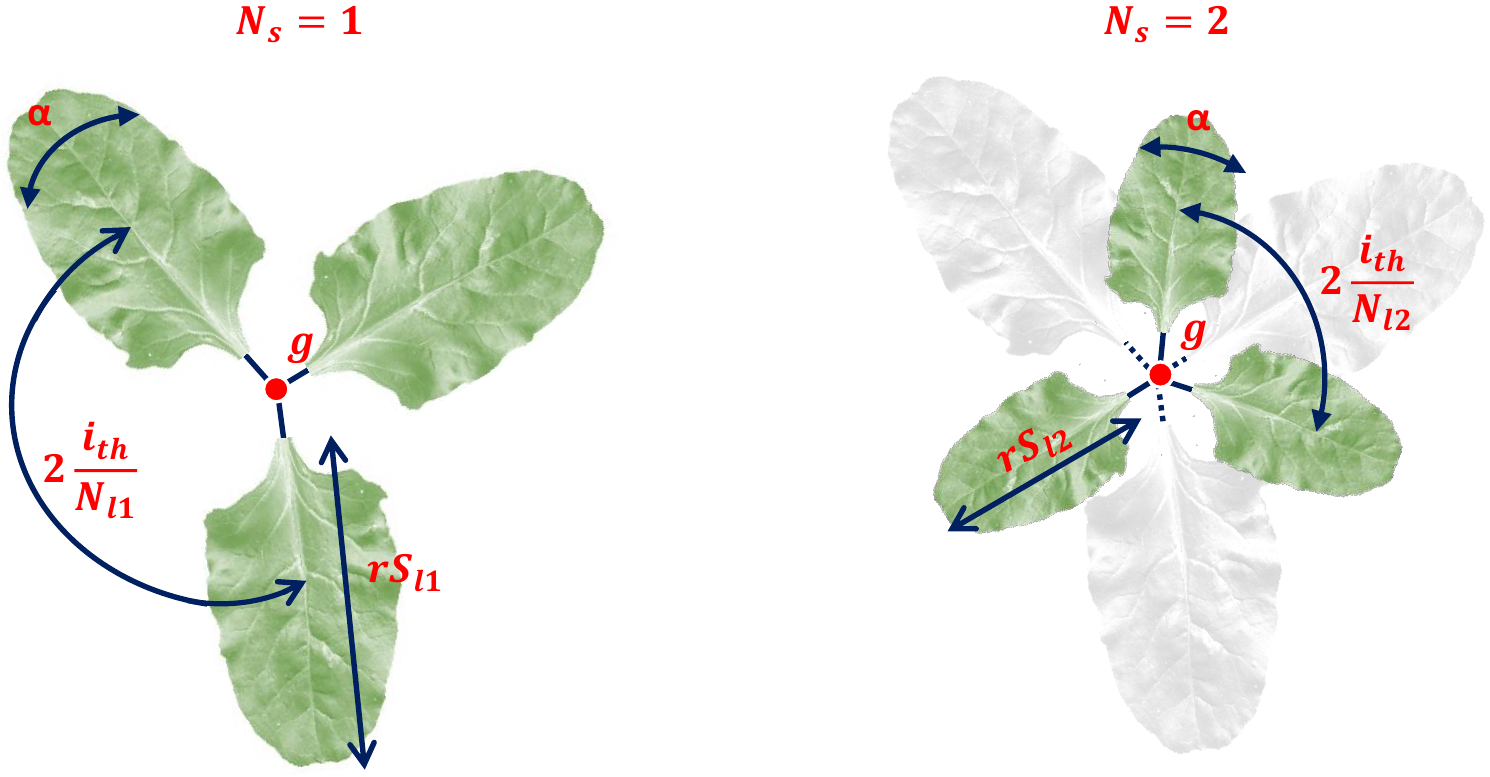}
		\caption{\small (a) Shows an example of a plant composed by a single layer leaves, while (b) represents the case of two layers arranged around the growth-axis \textit{g}.
		}
		\label{fig:multilayer_model}
	\end{figure}

	To enhance the rendering quality as well as the real world fidelity, we generate an approximation of the Normal Map (used to simulate high-resolution details on a low-resolution model, see Fig. \ref{fig:sugar_beet_leaf_process}(b)), an Ambient Occlusion map (used to approximate how bright light should be shining on any specific part of a surface, see Fig. \ref{fig:sugar_beet_leaf_process}(c)) and a Height Map (used to provide extra definition to the leaf model, see Fig. \ref{fig:sugar_beet_leaf_process}(d)).
	
	\subsection{Plant Modeling}
	\label{sec:plant_modeling}
		
	The modeling of an artificial plant follows a procedural scheme too. In order to resemble as much as possible the semblance of real plants, the synthetic plant model needs to take into account properties such as the average number of leaves per plant, their relative distribution, and the amount of leaf layers. We model the whole set of plant species with a generic multi-layer radial distribution of leaves. 
	
	According to fig. \ref{fig:multilayer_model}, an artificial plant is composed by three main entities: 
	\begin{itemize}
	
		\item \textit{growth stage axis}: it is the axis (called \textit{g} in Fig. \ref{fig:multilayer_model}) around which the leaves are placed. We parametrize such axis by the 3D position $p$ in the scene and the relative direction $d$ with respect to the gravity vector. 
		
		\item \textit{layers number}: the number of leaf layers $N_s$, that is usually related with the growth stage. 
		
		\item \textit{leaves per layer}: we distribute an average
number of $N_l$ leaves around the growth stage axis, where $N_l$ depends on the specific plant species. The i-th leaf is placed with an offset of $2\frac{i}{N_l}+\alpha$, where $\alpha$ represents a random component used to diversify the leaf arrangement. 
	
		\item \textit{size of leaves per layer}: as the previous parameters, the size $S_l$ is constrained to the age of the specimen. Conversely to the \textit{layers number}, we introduce a random variation on the leaf size by means of a multiplicative factor $r*S_l$ 
		
	\end{itemize}	
	
Among the above-mentioned parameters, we fix the number of leaves per layer $N_l$, the layers number $N_s$ and the size $S_l$, that depends on the plant species, choosing them accordingly to \cite{Meier_2001}. 

  In this way, modulating the remaining parameters $(p, d, \alpha, r)$, our system is able to procedurally generates a large variety of realistic crop and weeds instances. In order to give a further increment to the virtual environment variation, in each dataset we generate plants belonging to two different growth stages. 
  Since the initial growth stage is the most important in the targeted treatments, it is noteworthy to highlight how this model allows to cover most of the plant species we can found in a real agricultural environment. Examples of difference instance of virtual weeds obtained following our procedure are shown in Fig. \ref{fig:plant_specimans}.

\begin{figure}[ht]
	\centering
		\includegraphics[width=0.95\columnwidth]{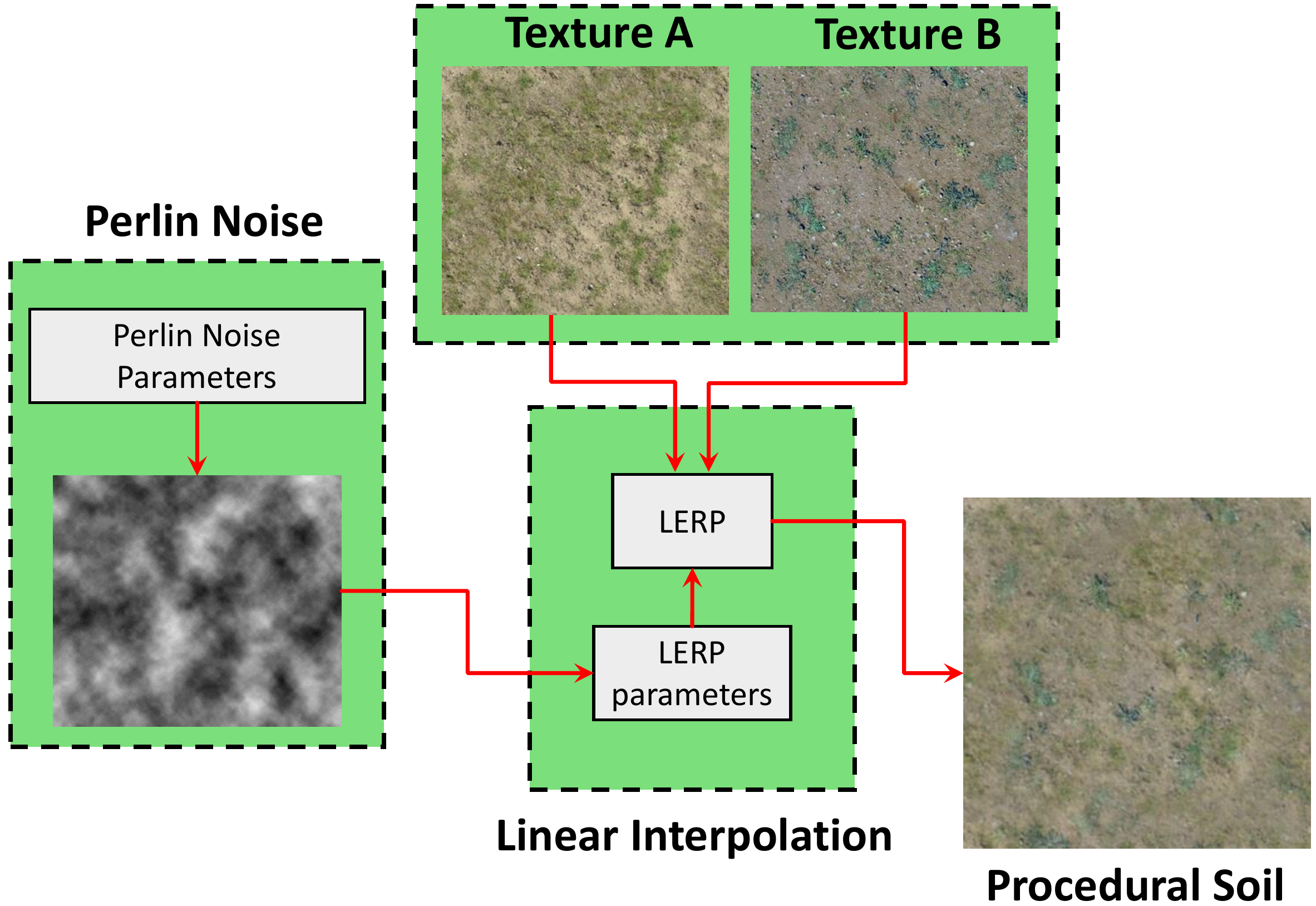}
		\caption{\small Example of terrain generation using two inputs. The image shows how two textures are blended together. A linear interpolation node has two textures and an intensity image as inputs. The intensity image is generated via Perlin Noise. The linear interpolation node blends the input images using the intensity input as selector.}
		\label{fig:terrain}
	\end{figure}	
	
\subsection{Virtual Dataset Generation}
	
		The final step required to render a realistic agricultural scene is represented by the virtual environment generation procedure, which consists of the following parts: soil generation, lightning and plants spawning.
		The ground is simulated as a simple planar surface, where from the fragments point of view we used different real world textures (see Fig.~\ref{fig:terrain}).
	 Each texture can represent different kinds of soil, i.e dirt, cracked dirt, stony..., and for each soil texture, we generate the Normal Map as well as the Ambient Occlusion and the Height Maps.
	These textures are blended together via Perlin Noise linear interpolation \cite{perlin1985image} to finally generate the soil, as shown in Fig. \ref{fig:terrain}. 	
	
	Following a similar blending procedure between Height Maps and the Normal Maps of the respective textures, we obtain a realistic vertices displacement.
	By modulating the Perlin Noise parameters, we can actually generate different configurations of the terrain both on the fragments and vertices point of view. 
	To resemble the light conditions of real world datasets, different light sources can be placed into the scene. In this way, the choice of the light source directly affects the scene illumination conditions, resulting in different shadowing behaviours.	
	
	Objects, such as plants, weeds, rocks, sticks.., are then spawned around the scene via a random distribution. 
	The normal direction of the soil surface sampled in the plant position is then used as initial growth stage direction corrupted by a small white noise perturbation. 
	The whole scene is finally projected into the image plane, using intrinsic and extrinsic camera parameters as close as possible to the real camera we are going to use for the actual crop/weeds detection.
	
	\subsection{Artificial Ground Truth Generation}
	
\begin{figure}[ht]
		\centering
		\begin{tabular}{c}
			\subfloat[RGB Image]{\includegraphics[width=0.9\columnwidth]{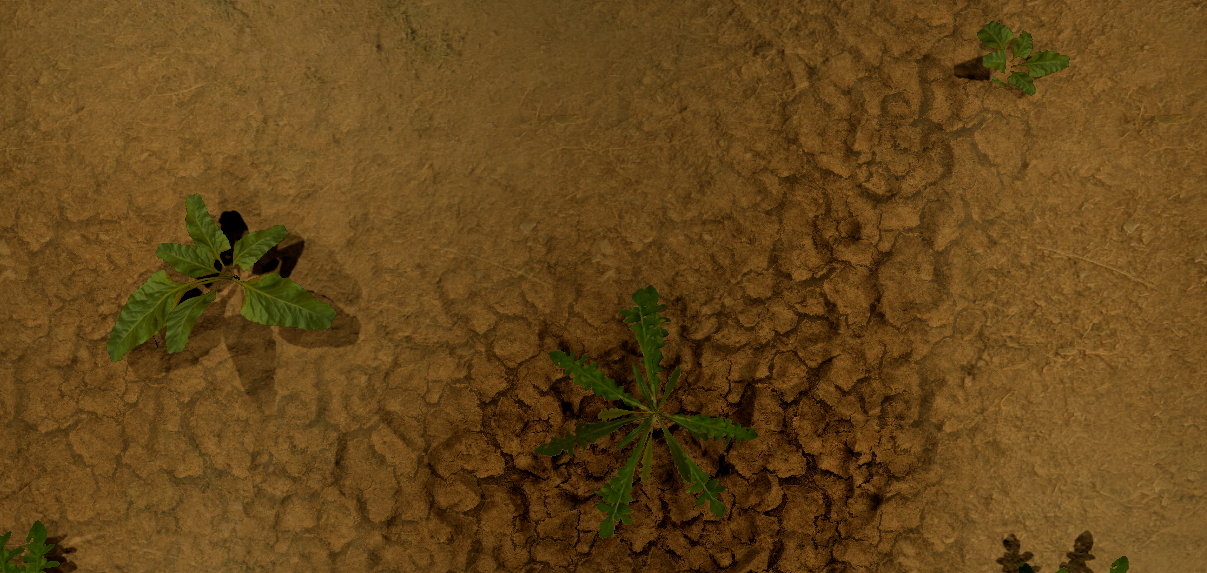}} \\
			\subfloat[Artificial Ground truth]{\includegraphics[width=0.9\columnwidth]{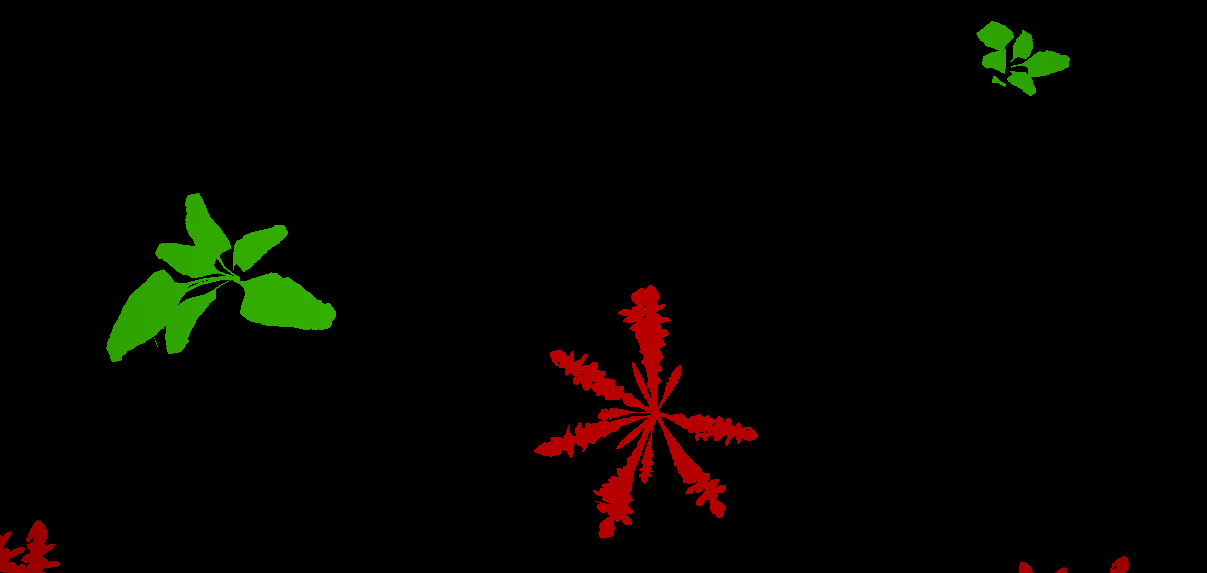}}\\
		\end{tabular}
		\caption{\small (a) Shows an example of a generated RGB image our system is capable to provide. In (b) each class have been labeled by using a different color. In the example, the system automatically highlights in green the objects that belong to the \textit{sugar beet} class, in red the \textit{weeds} (like the \textit{Capsella Bursa-Pastoris}) while the soil is turned to black. 
		}
		\label{fig:artificial_ground_truth}
	\end{figure}

	To generate the ground truth labeled images, we set the target object material as unlit and, as emissive component, we simply put the color of the belonging class. Thus, just turning off the anti-aliasing on the camera and all the lights in the scene, it is possible to easily obtain the annotation mask required to train the model. An example of a final rendered sample is shown in Fig. \ref{fig:artificial_ground_truth}, where  (a) is the artificial terrain RGB image and (b) is the synthetic ground truth.
	
\
	\section{Experiments}	
	
This experimental section is designed to support the main claim made in this paper: a synthetic dataset obtained by means of a procedural generative model and few real world textures can be used to train a modern machine learning framework, obtaining comparable results with respect to the same framework trained with a real dataset. This result enables a dramatic reduction of the human effort required to acquire and label real data. Furthermore, as the results suggest, a synthetically generated dataset can also be used to supplement small real datasets, enabling cutting-edge results. 

\subsection{Experimental Setup}

\begin{table*}[!ht]
		\centering
		\caption{ Image segmentation results.}
		\label{tab:classification}
		\begin{tabular}{ llllllll }
			\hline
			\multicolumn{2}{c}{} & \multicolumn{3}{c}{Test Set A} & \multicolumn{3}{c}{Test Set B} \\
			\cline{3-8}
			Net Variant & Train Set  & GA & CA & I/U & GA & CA & I/U\\\hline\hline
			\multirow{6}{*}{RGB SegNet} & Real A & 98.6  & 59.6 & 57.5 & 94.3 & 56.0 & 51.7 \\
			& Real B & 94.7 & 67.1 & 64.3  & 96.2 & 72.3 & 66.2  \\ 
			& Synthetic A & 99.2 & 66.8 & 47.4 & 96.3 & 74.9 & 55.4\\
			& Synthetic B & 97.8 & 62.2 & 53.3 & 95.5 & 60.2 & 55.1\\
			& Synthetic C & 98 & 61.3 & 53.9 & 96.1 & 59.7 & 55.3\\
			& Synthetic D & 97.6 & 62.5 & 53.1 & 95.6 & 60.1 & 55.2\\
			& Real-Augmented & 98.1 & 63.7 & 52.6 & 96.3 & 59.6 & 55.5\\\hline
			\multirow{6}{*}{RGB Basic SegNet} & Real A & \textbf{99.7} & \textbf{88.9}  & \textbf{80.3} & 97.5 & 88.4 & 72.4\\
			& Real B & 99.6 & 82.1 & 72.9 & 98.1 & 91.0 & \textbf{83.1}\\
			& Synthetic A & 99.4 &  80.3 & 58.9 & 96.4 & 80.9  & 58.3 \\
			& Synthetic B & 99.5  & 86.8  & 60.2 & 96.8 & 83.3 & 58.3 \\
			& Synthetic C & 99.5 & 70.2 & 55.1 & 96.5  & 55.9   & 52.5 \\
			& Synthetic D & 99.6 & 78.7  & 59.8 & 96.7 & 84.2 & 61.1 \\
			& Real-Augmented & 99.6 & 84.6  & 74.1 & \textbf{99.8}  & \textbf{91.3}  & 76.2\\\hline
		
		\end{tabular}
	\end{table*}	
	  
We use two datasets, both collected from a BOSCH Bonirob farm robot (Fig.~\ref{fig:teaser}) moving on a sugar beet field. Both the datasets are composed by a set of images taken by a 1296$\times$966 pixels 4-channels (RGB-NIR) JAI AD-130 camera mounted on the Bonirob. During the acquisition, the camera pointed downwards on the field and took images with a frequency of 1 Hz.\\
The first dataset (\textit{Real A}) is composed by 700 images and it has been collected in the first growth stage of the plants, when both crop and weeds have not yet developed their complete morphological features. The second dataset (\textit{Real B}) is composed by 900 images and it has been collected after 4 weeks: plants in this case are in an advanced growth stage. Each dataset has been manually labeled: the annotation procedure typically takes 5 to 30 minutes per image.

For the procedural dataset generation, we use Unreal Engine 4\footnote{Unreal Engine 4 is a complete open-source creation suite for game developers https://www.unrealengine.com/en-US/blog.} as graphic engine. Since we aim to make a direct comparison against the real-world data, we choose a comparable size for the generated synthetic datasets with respect to the real ones. 
We render four synthetic datasets, each one composed by 1300 images, using the method presented above:
(i) The first one (called \textit{Synthetic A} in the tables) includes only sugar beet plants and some random weeds.
(ii) \textit{Synthetic B} includes sugar beet plants and many instances of the \textit{Capsella Bursa-Pastoris} weed, that is the most common weed found in the real datasets. (iii) In \textit{Synthetic C} we include another kind of weed called \textit{Galium Aparine}. (iv) \textit{Synthetic D} contains sugar beet plants and all the aforementioned weed species. 
To address the case of a limited amount of annotated data, we also generated another dataset that we call \textit{Real-Augmented}. We compose this dataset by adding a random sample of 100 images from \textit{Real A} and 100 images from \textit{Real B} to \textit{Synthetic D}. 
Since we cannot simulate in a realistic manner the effect of the light reflection in the Near Infrared (NIR) channels, we used only the RGB data.
As test sets, we used two subsets removed from \textit{Real A} and \textit{Real B} (called \textit{Test Set A} and \textit{Test Set B} in the tables,  respectively). All the images have been resized to $480\times 360$ pixels. The performance have been measured by exploiting widely used metrics: global classification accuracy (acronym \textit{GA} in the tables) that provides the number of correct predictions divided by the number of all predictions; per-class average accuracy (\textit{CA}) that provides the mean over all classes of the number of correct predictions made for a specific class divided by the actual number of samples of this class;  average intersection over union (\textit{I/U}, see Eq.~\ref{eq:iou}); precision and recall (P and R, see Eq.~\ref{eq:pr}):

   \begin{equation}
      \textit{I/U} = \frac{1}{N} \sum_{i=1}^N \left( \frac{T_{pi} }{T_{pi} + F_{pi} + F_{ni}} \right) 
      \label{eq:iou}
   \end{equation}
   \begin{equation}
      \textit{P} =  \frac{T_{pi} }{T_{pi} + F_{pi}} , \textit{R} = \frac{T_{pi} }{T_{pi} + F_{ni}} 
      \label{eq:pr}
   \end{equation}

   where $i$ represents the class number, $N$ is the total number of classes, $T_{pi}$, $F_{pi}$ and $F_{ni}$ the number of true positives, false positives and false negatives, respectively, for the class $i$.  We finally multiplied all metrics by 100 to turn them into percentages.   
   
        \subsection{SegNet}	
	We evaluate the performance and quality of our synthetic databases using an effective pixel-wise classification CNN, \textit{SegNet} \cite{Kendall_2015}. \textit{SegNet} is composed by an encoder network and a corresponding decoder network, and finally by a pixel-wise softmax layer that performs pixel-wise classification.
	Differently from our previous work \cite{Potena_2016}, where the entire image is first divided into patches and then classified blob-wise with a voting scheme, \textit{SegNet} makes a pixel-wise semantic classification. This leads to a higher accuracy, since \textit{SegNet} is able to discriminate even in the case of overlapping plants. Such advantages make \textit{SegNet} particularly suitable for our purposes. 
	For the evaluation phase we chose to use \textit{SegNet} and a smaller version of this network, called \textit{Basic SegNet}. While the former is composed of 26 layers, the latter has only 8 layers. The main motivation that stands behind the choice of two different variants of the same network is the avoidance of overfitting. The bigger network has been developed for semantic segmentation of a large number of classes, prone to overfit in case of a small number of output classes. 

	\subsection{Crop/Weed Classification and Vegetation Detection}
	
	\begin{figure*}[!ht]
		\centering
		\begin{tabular}{cccc}
			{\includegraphics[width=0.22\textwidth]{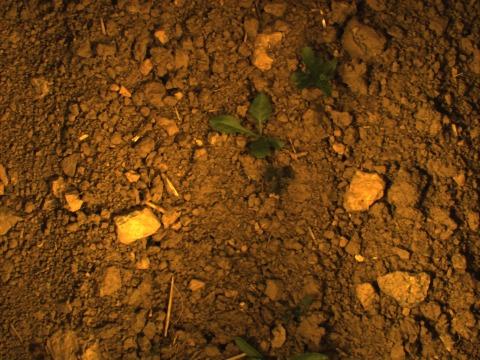}} & 
			{\includegraphics[width=0.22\textwidth]{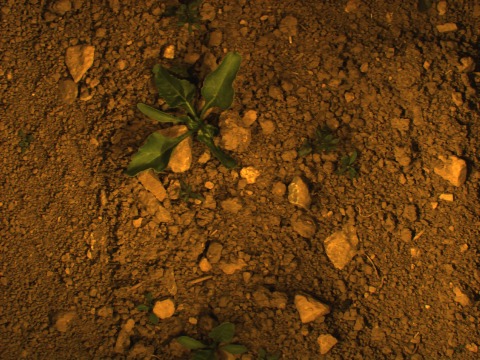}} &
			{\includegraphics[width=0.22\textwidth]{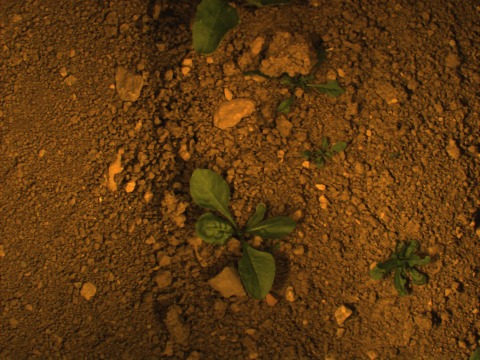}} &
			{\includegraphics[width=0.22\textwidth]{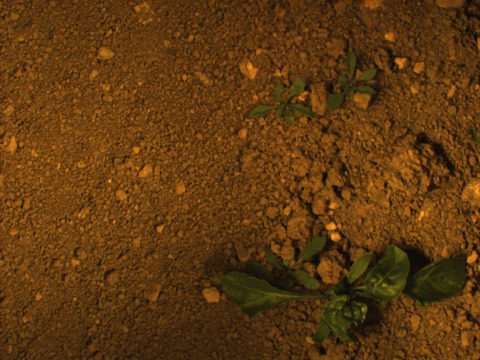}}\\
			{\includegraphics[width=0.22\textwidth]{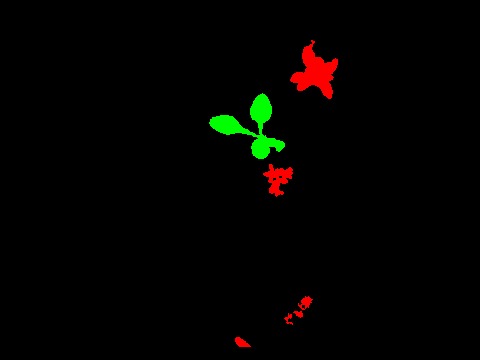}} & 
			{\includegraphics[width=0.22\textwidth]{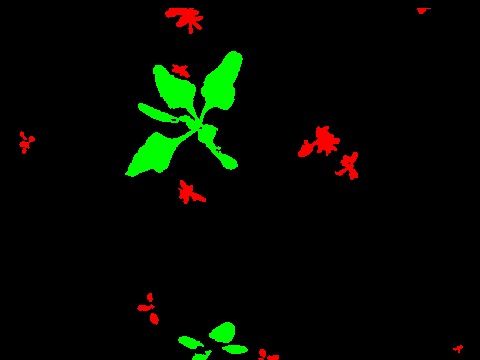}} &
			{\includegraphics[width=0.22\textwidth]{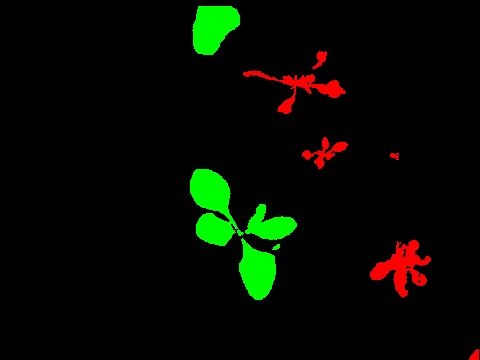}} &
			{\includegraphics[width=0.22\textwidth]{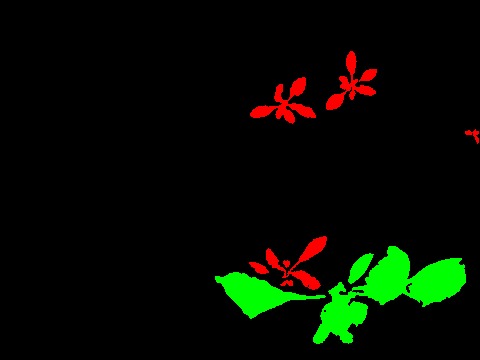}}\\
			{\includegraphics[width=0.22\textwidth]{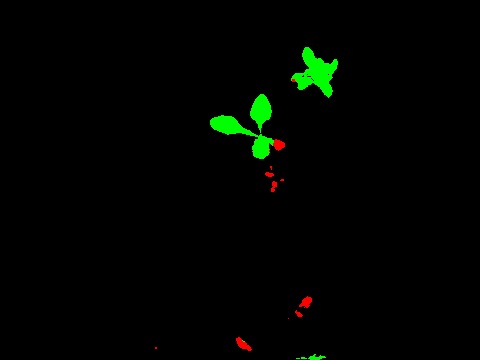}} & 
			{\includegraphics[width=0.22\textwidth]{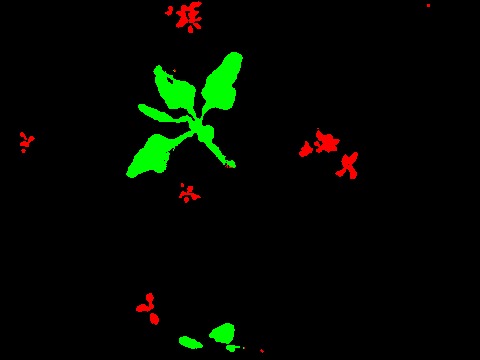}} &
			{\includegraphics[width=0.22\textwidth]{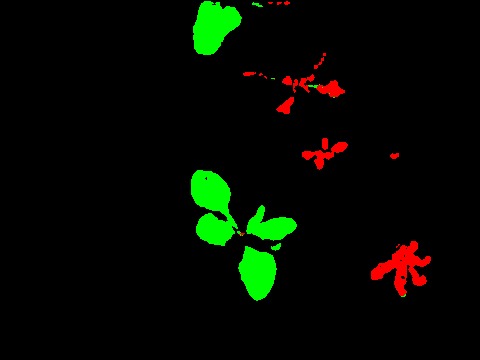}} &
			{\includegraphics[width=0.22\textwidth]{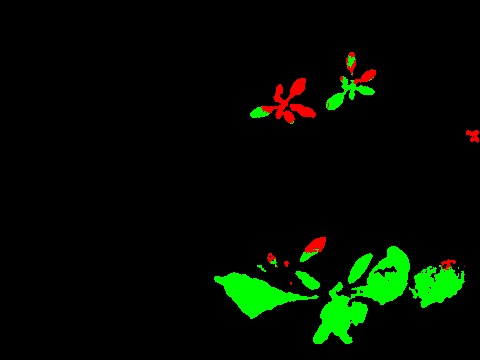}}\\            
		\end{tabular}
		
		\caption{\small Some examples of real images segmentation using an \textit{RGB Basic SegNet} trained using synthetic datasets. First row: input real RGB images. Second row: ground truth labels. Third row: segmentation results.\\ \\ \\}
		\label{fig:examples}
	\end{figure*}

The first set of experiments compare the performance among different crop/weeds classifiers. The difference is made in terms of training dataset and network typology, while the comparison is made by applying the trained models on both \textit{Test Set A} and \textit{Test Set B}.  We report the numerical results of the pixel-wise soil-crop-weeds classification in table \ref{tab:classification}, while in Fig.~\ref{fig:examples} we report some qualitative results. In this experiment, the most significant metrics are the per-class average accuracy and the average intersection over union. The global accuracy can be sometimes a misleading metric: for instance, in our test sets the majority of pixels represents soil in the scene, hence also a classifier that predicts only soil will still obtain good GA performance.\\
The best performance are generally obtained by the \textit{RGB Basic SegNet} network trained with real datasets (\textit{Real A} and \textit{Real B}) and with the \textit{Real-Augmented} dataset. Particularly, the latter often outperforms the real ones, achieving very good average intersection over union results for both \textit{Test Set A} and \textit{Test Set B} and a remarkable per-class average accuracy of 91.3\% on \textit{Test Set B}.

Moreover, also the results obtained by \textit{RGB Basic SegNet} trained \textit{solely} with the synthetic datasets are noteworthy, especially for the \textit{Synthetic B} and \textit{Synthetic D} datasets. These surprising results are also implicitly confirmed in \cite{Hattori_2015}, where the authors claimed that using synthetic data can outperform models trained on a limited amount of real scene-specific data.  Another interesting result is the significant performance difference between \textit{Synthetic B} and \textit{Synthetic C}. The main motivation that probably stands behind these fluctuating results is the actual weed distribution in the real-world datasets. Indeed, in such data the \textit{Capsella Bursa-Pastoris} plays the role of the most common weed while there are just a few instances of \textit{Galium Aparine}.

It is also important to note how the incremental inclusion of different weed species in the artificial datasets yields a monotone positive trend in the classification performance, converging toward the real-data performance when we also include real data in the training dataset. This fact implicitly confirms the goodness of our method.\\
On the other hand, the results obtained by the complete SegNet version \textit{RGB SegNet} are generally poor, also for real data, mainly due to overfitting phenomena. 

In Table \ref{tab:segmentation} we report the numerical results of the pixel-wise soil-vegetation classification. Also this evaluation confirms the previous results: also in this case the synthetically generated datasets are able to approximate the results obtained using the real datasets, while as before \textit{RGB Basic SegNet} performs much better in all cases compared with \textit{RGB SegNet}. \\
Here is noteworthy to highlight that we do not use the information coming from the NIR channel. Due to the high reflectance of green plants in this spectrum, this information is strictly related to the vegetation detection: we are investigating for a specific model to synthetically generate the NIR channel, aiding the soil-vegetation detection.

We finally report in Table \ref{tab:runtime} the infer time of each network using an NVidia GPU GTX 1070.

     \begin{table}[!ht]
     		\centering
     		\caption{ Vegetation detection results }
     		\label{tab:segmentation}
     		\begin{tabular}{ llllll }
     			\hline
     			\multicolumn{2}{c}{} & \multicolumn{2}{c}{Test Set A} & \multicolumn{2}{c}{Test Set B} \\
     			\cline{3-6}
     			Net Variant & Train Set  & P & R & P & R\\\hline\hline
     			\multirow{6}{*}{RGB SegNet} & Real A & 99.0 & 47.7 & 97.1 & 40.8\\
     			& Real B & 98.2  & 52.9 & 94.9  &  69.0 \\ 
     			& Synthetic A & 58.0 & 90.4& 53.8 & 49.3\\
     			& Synthetic B & 67.5 & 49.6 & 71.7 & 68.5\\
     			& Synthetic C & 63.2 & 64.2& 73.8 & 75.6\\
     			& Synthetic D & 67.5 & 67.6 & 77.9 & 76.9\\
     			& Real-Augmented & 68.1 & 65.1 & 83.3 & 84.2\\\hline
     			\multirow{6}{*}{RGB Basic SegNet} & Real A & 89.2 & 98.4 & 90.3 & 98.6\\
     			& Real B & 90.7 & 93.1 & 91.0 & 94.5\\
     			& Synthetic A & 80.9 & 90.9 & 78.5 & 86.2\\
     			& Synthetic B & 81.3 & 94.1 & 79.9 & 89.2\\
     			& Synthetic C & 77.6 & 91.4 & 69.2 & 87.1\\
     			& Synthetic D & 82.2 & 94.0 & 80.4 & 89.7\\
     			& Real-Augmented & 83.1 & 94.2 & 83.6 & 92.8\\\hline
     		\end{tabular}
     	\end{table}

	\begin{table}[!ht]
		\centering
		\caption{ Infer runtime}
		\label{tab:runtime}
		\begin{tabular}{ll}
			\hline\noalign{\smallskip}
			Net Variant & time (s)\\   
			\noalign{\smallskip}
			\hline
			\noalign{\smallskip}
			RGB SegNet & 0.14 \\
			RGB Basic SegNet &  0.08 \\
			\hline
		\end{tabular}
	\end{table}			
	
%
%
%
%
	
	\section{Conclusions}
    In this paper, we presented an approach to minimize the human effort required to train a visual model classification system.
    We make an explicit modeling of the target environment and, using few real world textures, we generate a large variety of annotated data of plant specimens using a procedural scheme. 
      
    By synthetically generating training datasets we removed the human intervention in the labeling phase, reducing the time and the effort needed to train a visual inference model.
    The system trained using synthetic data shows a similar performance than the one trained with real-world human annotated images.
	The artificial data can be used even in presence of a real dataset with a limited amount of data, just as supplement.    
	Our results look promising and also suggest that a virtual dataset containing all the weeds we can find in a real field can lead to the same level of accuracy of real-world images.   
    As a future work, we are going to investigate for a chlorophyll model in order to render also realistic NIR images, a common and useful source of data in the precision agriculture context, improving the overall performance of the system.\\

\balance
	\bibliographystyle{IEEEtran}
	\bibliography{IEEEabrv,crop}

\end{document}